\documentclass[runningheads]{llncs}
\usepackage{graphicx}
\usepackage{amsmath}
\usepackage{algorithm}
\usepackage{algorithmic}
\usepackage{amssymb}
\usepackage{subfig}
\usepackage{xcolor}
\usepackage{arydshln}
\usepackage{multirow}
\usepackage{orcidlink}
\usepackage{marvosym} 
\usepackage{ifsym}
\usepackage{bm} % for bold gamma

\DeclareMathAlphabet\mathbfcal{OMS}{cmsy}{b}{n}
\def\0{{\bf 0}}
\def\1{{\bf 1}}

\usepackage{pifont} % for √ and ×

\newcommand{\topline}{\hline}

\newcommand{\doublemidline}{\hline\hline}

\begin{document}
\title{Sensitivity-Aware Post-Training Quantization for Deep Neural Networks}
%
%\titlerunning{Abbreviated paper title}
% If the paper title is too long for the running head, you can set
% an abbreviated paper title here
%
\author{Zekang Zheng\inst{1} \and Haokun Li\inst{1} \and Yaofo Chen\inst{1(\textrm{\Letter})}\orcidID{0000-0001-8916-8368} \and Mingkui Tan\inst{1,2,3}\orcidID{0000-0001-8856-756X} \and Qing Du\inst{1}}

\institute{South China University of Technology, Guangzhou, China \\
\email{\{chenyaofo\}@scut.edu.cn} \and
Pazhou Laboratory, Guangzhou, China \and
Key Laboratory of Big Data and Intelligent Robot, Ministry of Education}
\maketitle              % typeset the header of the contribution
\begin{abstract}
    Model quantization reduces neural network parameter precision to achieve compression, but often compromises accuracy. Existing post-training quantization (PTQ) methods employ iterative parameter updates to preserve accuracy under high compression ratios, incurring significant computational complexity and resource overhead, which limits applicability in resource-constrained edge computing and real-time inference scenarios. This paper proposes an efficient PTQ method guided by parameter sensitivity analysis. The approach prioritizes quantization of high-sensitivity parameters, leveraging unquantized low-sensitivity parameters to compensate for quantization errors, thereby mitigating accuracy degradation. Furthermore, by exploiting column-wise clustering of parameter sensitivity, the method introduces a row-parallel quantization framework with a globally shared inverse Hessian matrix update mechanism, reducing computational complexity by an order of magnitude. Experimental results on ResNet-50 and YOLOv5s demonstrate a 20--200-fold quantization speedup over the Optimal Brain Quantization baseline, with mean accuracy loss below 0.3\%, confirming the method’s efficacy in balancing efficiency and accuracy.

\end{abstract}
    
\section{Introduction}
The widespread application of deep learning models in domains such as computer vision~\cite{he2016deep,liu2021swin} and natural language processing~\cite{brown2020gpt3,openai2023gpt4,touvron2023llama,liu2024deepseek} has led to a continual increase in their scale and computational complexity. This poses significant challenges for deploying these models on resource-constrained devices. Model quantization, a crucial model compression technique, reduces the numerical precision of model parameters and activations. This not only significantly diminishes model storage requirements and computational overhead but also enables efficient model deployment and inference. However, the quantization process, being inherently an approximation, inevitably leads to performance degradation. Consequently, achieving efficient and rapid quantization while preserving model accuracy has become a prominent research focus of model compression.

Model quantization methods are generally categorized into two types: Post-Training Quantization (PTQ) and Quantization-Aware Training (QAT). While QAT methods can achieve higher quantization accuracy, they complicate and prolong the training process due to the introduction of quantization operations during training. In contrast, PTQ methods offer greater flexibility, obviating the need for model retraining and allowing direct quantization of pre-trained models, which facilitates deployment in practical applications. Nevertheless, PTQ methods often encounter the issue of post-quantization accuracy degradation. Thus, designing PTQ algorithms that are both efficient and maintain high accuracy presents a significant contemporary challenge. Most existing PTQ methods are based on a layer-wise quantization approach, where each layer is quantized independently. For instance, AdaRound~\cite{nagel2020up} employs gradient optimization to learn rounding strategies, while the classical Optimal Brain Surgeon (OBS) framework~\cite{lecun1989optimal} utilizes Taylor expansion to minimize layer-wise quantization error. However, these methods still exhibit limitations such as considerable accuracy loss, high computational overhead, and extended quantization times. Particularly in edge computing scenarios, where model updates are frequent, the efficiency of the quantization process is paramount. Although methods like Optimal Brain Quantizer (OBQ)~\cite{frantar2022optimal} demonstrate good accuracy, their row-wise parameter update scheme limits parallelism, leading to prolonged quantization times and rendering them unsuitable for the real-time demands of edge environments.

To address the aforementioned challenges, this paper proposes a sensitivity-guided efficient post-training quantization algorithm. The core idea is to first pre-sort the columns of the parameter matrix and subsequently perform column-wise quantization based on this sorted order. To circumvent the problem of repetitive Hessian matrix computations encountered in existing methods, we introduce a parallel parameter quantization algorithm operating along the row dimension. This design allows all rows to share and update a single Hessian matrix (or its inverse, more specifically), significantly reducing computational overhead. Furthermore, we observe that values exhibiting large quantization errors within the model often follow a long-tail distribution. Based on this observation, we propose a column-wise quantization algorithm under the pre-sorted parameter matrix. By prioritizing the quantization of columns associated with larger errors, we can more effectively leverage the compensation mechanism inherent in OBS-like algorithms~\cite{lecun1989optimal}, thereby enhancing overall quantization accuracy. By minimizing unnecessary computations and prioritizing more sensitive parameters, our method achieves an effective balance between quantization efficiency and accuracy. The proposed method requires no retraining and completes the quantization process in a relatively short time, making it suitable for scenarios with stringent real-time requirements, such as edge computing.

\noindent\textbf{Contributions}:
1) We designed a sensitivity-guided model parameter quantization strategy, which effectively mitigates accuracy loss during the quantization process.
2) We proposed a parallel parameter quantization algorithm along the row dimension. By establishing a shared inverse Hessian update mechanism, parameter quantization operations are executed in parallel along the row dimension, significantly reducing computational overhead and substantially decreasing quantization time.
3) Extensive experimental results show that our method significantly reduces quantization time and memory footprint across various models, achieving nearly lossless accuracy compared to current SoTA methods.

\section{Related Work}
Model quantization, a critical technique for neural network compression, aims to map weights and activations, typically represented as floating-point numbers, to lower-precision discrete values (e.g., fixed-point integers), thereby significantly reducing memory footprint, computational latency, and power consumption~\cite{gholami2022survey}. This technique is vital for deploying increasingly large and complex deep learning models in resource-constrained environments, such as edge devices and mobile platforms. Empirical evidence shows that converting floating-point representations to 4-bit or 8-bit fixed-point integers can achieve 4x to 8x reductions in memory and latency, with theoretical reductions up to 16x, while striving to preserve the model's original accuracy.

\textbf{Post-Training Quantization (PTQ)}~\cite{banner2019post,choukroun2019low,hubara2021accurate,li2021brecq,liu2021post,nagel2020up} offers an efficient method to convert pre-trained floating-point models to low-precision formats. PTQ primarily involves quantizing the weights and activations of a trained model, providing significant advantages for rapid deployment on resource constrained devices by reducing memory usage and computational demands. A common PTQ strategy is to compute the minimum and maximum values of model parameters to determine clipping ranges~\cite{hubara2021accurate}. This approach effectively reduces the data representation range while maintaining acceptable accuracy. Additionally, a small set of unlabeled data can be used as a calibration set~\cite{he2018learning,hubara2021accurate} to fine-tune quantization parameters. In practice, PTQ typically maintains good model performance at higher bit widths (e.g., 8 bits) with minimal accuracy loss. However, significant performance degradation may occur at lower bit widths. To address accuracy loss in PTQ, researchers have proposed several strategies, including correcting biases in the mean and variance of quantized weights~\cite{banner2019post,meller2019same} to better align with the model's requirements, balancing weight ranges across layers or channels to reduce cumulative quantization errors, and optimizing the L2 distance between quantized and floating-point tensors~\cite{choukroun2019low}. Techniques such as outlier channel splitting~\cite{zhao2019improving} also mitigate the impact of anomalous channels. Recent advancements, such as output reconstruction-based PTQ methods like AdaRound~\cite{nagel2020up} and AdaQuant~\cite{hubara2021accurate}, have emerged as prominent trends, enhancing PTQ accuracy and stability through output optimization.

\textbf{Quantization-Aware Training (QAT)}~\cite{jacob2018quantization,wang2019learning,nagel2022overcoming,liu2023llm} introduces pseudo-quantization operations during the training or fine-tuning phase, simulating the effects of quantization on forward propagation and gradient computation. QAT involves retraining on the training dataset using gradient descent to adjust model parameters, mitigating accuracy loss due to quantization. Although QAT requires more computational resources and time compared to PTQ, it significantly improves prediction accuracy at lower quantization bit widths, making it indispensable for high-accuracy applications. During QAT, floating-point computations are used for forward and backward propagation to ensure accuracy, while pseudo-quantization operators are inserted after gradient updates to simulate quantization effects. Handling non-differentiable operations, such as rounding and clipping, during backpropagation is a core challenge. To address this, researchers have proposed solutions like designing quantization operations with smooth, non-zero derivatives (e.g., LQ-Nets~\cite{zhang2018lq} and DSQ~\cite{gong2019differentiable}). However, the most widely adopted approach is the Straight-Through Estimator (STE)~\cite{bengio2013estimating}, which approximates non-differentiable operators with an identity function. Despite its simplicity, STE has proven effective in numerous experiments and applications, remaining a cornerstone of QAT.

\section{Problem Definition and Background}

\textbf{Layer-wise Compression Problem.} 
Let $f_\ell(X_\ell, W_\ell)$ be a layer,
where $W_\ell$ represents the weights and $X_\ell$ denotes the input. The goal of layer-wise compression is to find a compressed version of the weights, $\hat{W}_\ell$, that performs as closely as possible to the original weights $W_\ell$. Specifically, the compressed weights $\hat{W}_\ell$ should minimize the expected change in layer output, as measured by a loss function $\mathcal{L}$, while satisfying a general compression constraint $C(\hat{W}_\ell) > C$, where $C(\cdot)$ is tailored to the type of compression:
\begin{equation}
\arg\min_{\hat{W}_\ell}  E \left( f_\ell(X_\ell, W_\ell), f_\ell(X_\ell, \hat{W}_\ell) \right), \quad \text{subject to} \quad C(\hat{W}_\ell) > C.
\end{equation}

The weights $W_\ell$ form a $d_{\text{row}} \times d_{\text{col}}$ matrix (for convolutional layers, $d_{\text{col}}$ is the total number of weights in a single filter), the input $X_\ell$ has dimensions $d_{\text{col}} \times N$, the function $f_\ell(X_\ell, W_\ell)$ is defined as $W_\ell X_\ell$, and the error function $E$ is defined as $\|f_\ell(X_\ell, W_\ell) - f_\ell(X_\ell, \hat{W}_\ell)\|_2^2$. Thus, the quantization objective becomes:
\begin{equation}
\arg\min_{\hat{W}_\ell} \|W_\ell X_\ell - \hat{W}_\ell X_\ell\|_2^2, \quad \text{subject to} \quad C(\hat{W}_\ell) > C.
\end{equation}

\textbf{Optimal Brain Surgeon Method~\cite{dong2017learning}.} By approximating the target function $E$ using a Taylor series expansion, a perturbation $\delta \mathbf{w}$ in the parameter vector induces the following change in the error function. For a neural network, the change in error can be expressed via Taylor expansion as:
\begin{equation}
\delta E = \left( \frac{\partial E}{\partial \mathbf{w}} \right)^T \cdot \delta \mathbf{w} + \frac{1}{2} \delta \mathbf{w}^T \cdot \mathbf{H} \cdot \delta \mathbf{w} + \mathcal{O}(\|\delta \mathbf{w}\|^3),
\end{equation}
where $\frac{\partial E}{\partial \mathbf{w}}$ is the first-order derivative (gradient) of the error function $E$ with respect to the weights $\mathbf{w}$, $\mathbf{H}$ is the Hessian matrix containing all second-order derivatives of $E$ with respect to $\mathbf{w}$, $\delta \mathbf{w}$ represents the weight perturbation, and $\mathcal{O}(\|\delta \mathbf{w}\|^3)$ denotes higher-order terms that are neglected. For a neural network trained to a local minimum, the first-order derivative is zero, so the first term is omitted; higher-order terms are also neglected, leaving only the second-order term. Thus, the Taylor expansion simplifies to:
\begin{equation}
\delta E = \frac{1}{2} \delta \mathbf{w}^T \cdot \mathbf{H} \cdot \delta \mathbf{w}.
\end{equation}

To minimize the error increase after setting a weight to zero, the following constraint is introduced:
\begin{equation}
\mathbf{e}_q^T \cdot \delta \mathbf{w} + w_q = 0,
\end{equation}
where $\mathbf{e}_q$ is a unit vector in the weight space corresponding to the scalar weight $w_q$, which is the weight to be set to zero. Thus, the optimization objective becomes:
\begin{equation}
\min_q \left\{ \min_{\delta \mathbf{w}} \left( \frac{1}{2} \delta \mathbf{w}^T \cdot \mathbf{H} \cdot \delta \mathbf{w} \right) \ \middle| \ \mathbf{e}_q^T \cdot \delta \mathbf{w} + w_q = 0 \right\}.
\end{equation}

To solve this optimization problem, a Lagrangian function is constructed:
\begin{equation}
\mathcal{L} = \frac{1}{2} \delta \mathbf{w}^T \cdot \mathbf{H} \cdot \delta \mathbf{w} + \lambda \left( \mathbf{e}_q^T \cdot \delta \mathbf{w} + w_q \right),
\label{eq:optimize_goal}
\end{equation}
where $\lambda$ is the Lagrange multiplier. By taking derivatives with respect to $\delta \mathbf{w}$ and $\lambda$ and combining with the constraint, the following solution is obtained:
\begin{equation}
\delta \mathbf{w} = -\frac{w_q}{[\mathbf{H}^{-1}]_{qq}} \mathbf{H}^{-1} \cdot \mathbf{e}_q,
\end{equation}
and the error increase after setting the weight to zero is:
\begin{equation}
L_q = \frac{w_q^2}{2 [\mathbf{H}^{-1}]_{qq}}.
\end{equation}

Here, $L_q$ is referred to as the sensitivity of weight $q$, representing the error increase introduced by quantizing that weight. A more sensitive weight results in a larger error after quantization. Thus, the quantization process focuses on minimizing the growth of $L_q$ while quantizing all parameters.

\textbf{The Optimal Brain Quantizer Method~\cite{frantar2022optimal}.}This method is extended to the quantization domain. Suppose the weights of a layer are to be quantized on a fixed grid with width $\Delta$ while minimizing the loss. To map OBS to quantization, the Lagrangian constraint in Eqn.~\ref{eq:optimize_goal} is set to $(\text{quant}(w_p) - w_p)$, where $\text{quant}(w_p)$ is the quantized weight after rounding. 

With the update for the remaining parameters given by:
\begin{equation}
\delta w_p = -\frac{w_p - \text{quant}(w_p)}{[\mathbf{H}^{-1}]_{pp}} \cdot \mathbf{H}^{-1}_{:,p}.
\label{eq:quan_deltaw}
\end{equation}
The introduced quantization error is:
\begin{equation}
L_q = \frac{(\text{quant}(w_p) - w_p)^2}{2 [\mathbf{H}^{-1}]_{qq}}.
\end{equation}

\section{Sensitivity Guided Post-Training Quantization}

\subsection{Parameter Quantization Guided by Sensitivity Ranking}

\begin{figure}[t]
    \centering
    \includegraphics[width=1.0\linewidth]{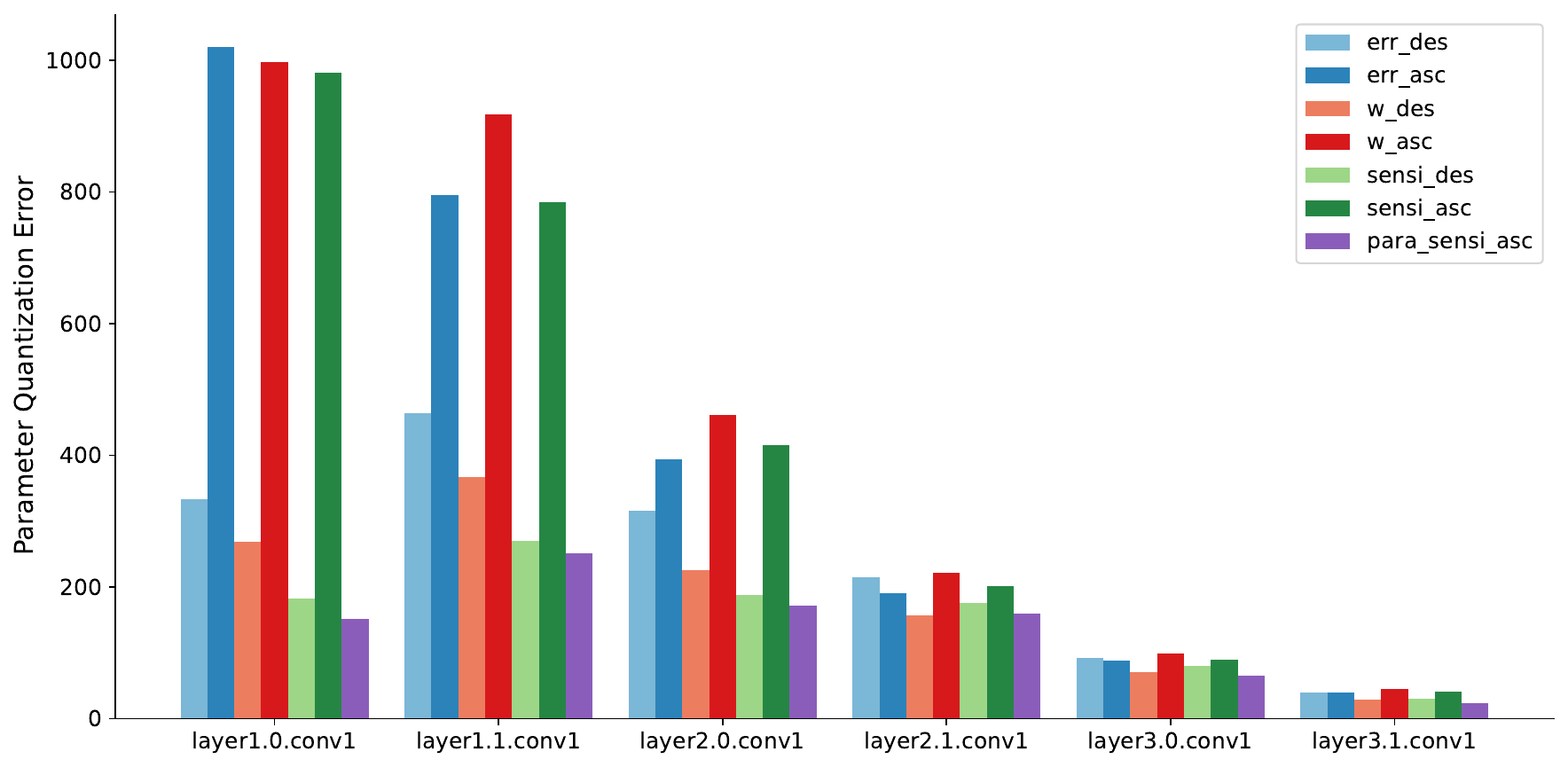}
    \caption{Quantization error under various sorting strategies.}
    \label{fastobq:loss}
\end{figure}

Adapting model pruning techniques, such as quantizing parameters row-wise in ascending order of pre-computed sensitivity $L_q$, is suboptimal for quantization. Pruning removes low-sensitivity parameters with minimal impact, but quantization processes all parameters within a layer. Quantizing in ascending $L_q$ order, which prioritizes low-sensitivity parameters, defers critical high-sensitivity parameters to later stages. In Figure~\ref{fastobq:loss}, this increases quantization error, as fixed low-sensitivity parameters limit compensation for errors from high-sensitivity ones. The figure uses colors to represent quantization orders, with prefixes ($err$, $w$, $sensi$) denoting sorting by quantization error, parameter norm, or sensitivity $L_q$, and suffixes ($des$, $asc$) indicating descending or ascending order.

To address this, we propose a sensitivity-guided quantization strategy that prioritizes parameters with the highest $L_q$. Quantizing high-sensitivity parameters first introduces primary errors early, when many unquantized low-sensitivity parameters remain. Per Eqn.~\ref{eq:quan_deltaw}, these low-sensitivity parameters collectively compensate for larger errors, unlike ascending quantization, which introduces errors late with limited adjustment capacity. Figure~\ref{fastobq:loss} shows that descending $L_q$ sorting significantly reduces layer-wise quantization error compared to other strategies, with $L_q$-based descending order performing best.

The proposed algorithm operates as follows: for a layer, compute $L_q$ for all parameters using Taylor expansion, sort them in descending $L_q$ order, and quantize sequentially. This approach mitigates the limitations of ascending quantization by leveraging low-sensitivity parameters for error compensation, achieving lower global quantization error and better model performance.

\begin{figure}[!ht]
    \centering
    \includegraphics[width=1\linewidth,height=10.4cm]{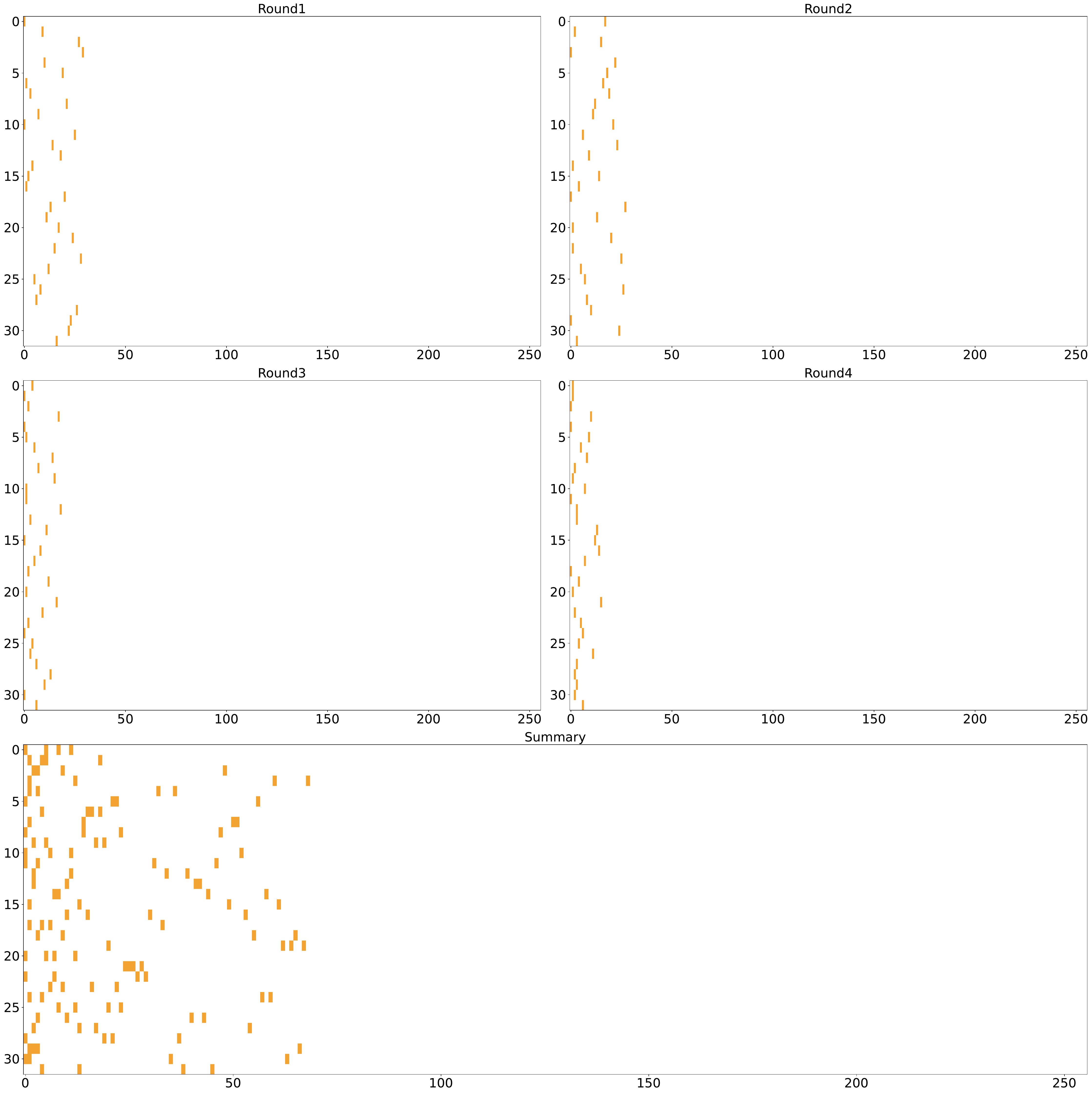}\\
    \caption{Row-wise quantization positions on ResNet18 (layer 1.0.conv1).}
    \label{fastobqL:ResNet-vis}
\end{figure}

\subsection{Row-Parallel Efficient Quantization Method}

Building on the analysis above, we design a quantization strategy inspired by Optimal Brain Quantization (OBQ). It processes weight matrix $W$ row-wise, quantizing parameters based on Hessian-derived sensitivity. After quantizing $w_{ij}$ to $q(w_{ij})$, OBQ updates unquantized parameters and the inverse Hessian $H^{-1}$:
\begin{equation}
    W \leftarrow W - H^{-1} \cdot \frac{1}{[H^{-1}]_{kk}} \cdot \delta_k
    \label{eq:update_w_cn}
\end{equation}
\begin{equation}
    H^{-1} \leftarrow H^{-1} - \frac{1}{[H^{-1}]_{kk}} H^{-1}_{:,k} H^{-1}_{k,:}
    \label{eq:update_H_cn}
\end{equation}
where $\delta_k$ is the quantization error vector (non-zero at index $k$ for $w_{ij}$).

From Eqn.~\ref{eq:update_H_cn}, the $H^{-1}$ update depends on index $k$ (position $(i, j)$). In OBQ, quantizing low-sensitivity parameters independently per row leads to unsynchronized column indices $j$ across rows $i$, requiring a distinct $H^{-1}$ for each row.

This requires storing a $[d_{\text{row}}, d_{\text{col}}, d_{\text{col}}]$ tensor for $H^{-1}$ matrices, with memory scaling quadratically with $d_{\text{col}}$ and linearly with $d_{\text{row}}$. Sequential row-specific updates hinder parallelization, increasing time complexity. A strategy quantizing the same column across rows could use a single $H^{-1}$, reducing overhead.

Based on sensitivity-guided ranking, we analyzed parameter sensitivity distribution. In Figure~\ref{fastobqL:ResNet-vis}, we visualize a ResNet-18 convolutional layer, showing high-sensitivity parameter columns in early rounds (truncated to 256 columns). The results show column-wise clustering, suggesting synchronized quantization is feasible without altering sensitivity ranking.

In Figure~\ref{fastobq:loss}, we present the per-layer error (``para\_sensi\_des''), comparable to the non-parallel ``sensi\_des'' method, validating that column-wise parallel quantization leverages sensitivity clustering effectively with minimal error increase.

Based on these observations and validations, we propose an efficient post-training quantization algorithm guided by sensitivity. The core idea shifts from parameter-level to column-level processing, guided by aggregated sensitivity. Specifically, given a weight matrix $W$ to be quantized:
\begin{enumerate}
    \item Compute the sensitivity $L_q$ for all individual parameters $w_{ij}$, typically derived from the diagonal elements of the inverse Hessian matrix $H^{-1}$.
    \item Instead of sorting individual parameters, compute an aggregated sensitivity score for each \textit{column} $j$ by summing the sensitivities $L_q$ of all parameters $w_{ij}$ in that column: $S_j = \sum_{i=1}^{d_{\text{row}}} L_q(w_{ij})$.
    \item Sort the columns of the weight matrix $W$ based on these aggregated sensitivity scores $S_j$ in descending order, prioritizing columns with higher total sensitivity. Accordingly, permute the rows and columns of the single $H^{-1}$ matrix to match the new column order.
    \item Perform quantization column-wise in the determined order. In each step, all parameters $w_{ij}$ in the current column $j$ (across all rows $i=1, \dots, d_{\text{row}}$) are quantized simultaneously.
    \item After quantizing an entire column $j$, perform a single update step, similar to Eqns.~\ref{eq:update_w_cn} and~\ref{eq:update_H_cn}, to reflect the collective impact of quantizing that column. This single update modifies the remaining unquantized parameters and the shared inverse Hessian matrix $H^{-1}$.
\end{enumerate}

Let the computational cost associated with a single inverse Hessian update (Eqn~\ref{eq:update_H_cn}) be denoted as $f(H^{-1})$. The original OBQ method, due to its row-wise processing of potentially different column indices, requires approximately $d_{\text{row}} \times f(H^{-1})$ computations for Hessian updates after processing one parameter per row. In contrast, the proposed column-wise method performs only one $f(H^{-1})$ operation per column quantization. This consolidation of updates significantly reduces the computational complexity of Hessian management, potentially by a factor proportional to $d_{\text{row}}$. Additionally, the need to store and manage only a single inverse Hessian matrix greatly reduces the memory footprint.

\section{Algorithm Complexity Comparison and Analysis}

The FastOBQ algorithm, is analyzed here for a single layer. The complexity analysis consists of the following components: computing parameter sensitivity, sorting parameters by column, parameter quantization and updates, and inverse Hessian matrix updates. Assume the parameters to be quantized form a $d_{\text{row}} \times d_{\text{col}}$ matrix. The computational complexity of each component is as follows:

\begin{enumerate}
    \item Computing Parameter Sensitivity: This involves calculating the inverse of the Hessian matrix. In practice, this is achieved by first performing Cholesky decomposition, with a time complexity of $O(d_{\text{col}}^3)$. Computing the inverse of the decomposed result also has a time complexity of $O(d_{\text{col}}^3)$. Thus, the total time complexity for this step is $O(d_{\text{col}}^3)$.
    \item Sorting Parameters by Column: For a matrix with $d_{\text{col}}$ columns, the sorting time complexity is $O(d_{\text{col}} \log d_{\text{col}})$.
    \item Parameter Quantization: Quantization is a linear operation, and parallelization across rows results in a time complexity of $O(d_{\text{col}})$.
    \item Parameter Updates: This involves element-wise vector multiplication, also parallelized, with a computational complexity of $O(d_{\text{row}} d_{\text{col}})$.
    \item Inverse Hessian Matrix Updates: This involves matrix multiplication of dimensions $[d_{\text{col}}, 1]$ and $[1, d_{\text{col}}]$, yielding a computational complexity of $O(d_{\text{col}}^2)$.
\end{enumerate}

Since steps (2), (3), (4), and (5) are performed sequentially for each column, the overall complexity of the FastOBQ algorithm is:
\[
O(d_{\text{col}}^3) + O(d_{\text{col}}) \times \left( O(d_{\text{col}} \log d_{\text{col}}) + O(d_{\text{col}}) + O(d_{\text{row}} d_{\text{col}}) + O(d_{\text{col}}^2) \right),
\]
which simplifies to $O(d_{\text{col}}^3) + O(d_{\text{row}} d_{\text{col}}^2)$. Compared to the OBQ's complexity of $O(d_{\text{row}} d_{\text{col}}^3)$, FastOBQ reduces the complexity by an order of magnitude.

\section{Experiments}

\subsection{Experimental Setup}
  
\noindent\textbf{Datasets}.
We use ImageNet~\cite{deng2009imagenet} and COCO~\cite{lin2014microsoft} to validate the proposed row-parallel post-training quantization algorithm under sensitivity-guided ordering. ImageNet (ILSVRC2012) contains 1.28 million training and 50,000 validation images across 1,000 categories, ideal for image classification. COCO, with 118,000 training and 5,000 validation images annotated for 80 object categories, supports object detection and segmentation tasks.

\noindent\textbf{Models}.
We employ ResNet~\cite{he2016deep} series (ResNet-18, 11.7M parameters; ResNet-34, 21.8M; ResNet-50, 25.6M) for classification, leveraging their residual structures to assess quantization sensitivity across depths. For detection, YOLOv5 variants (YOLOv5s, 7.2M parameters; YOLOv5m, 21.2M) are used, adjusting complexity via width and depth to evaluate multi-scale feature quantization.

\noindent\textbf{Evaluation Metrics}.
For ImageNet classification, Top-1 Accuracy measures the proportion of correct predictions. For COCO detection, mean Average Precision (mAP) is used, with mAP@0.5 at an IoU threshold of 0.5 and mAP@0.5:0.95 averaged over IoU thresholds from 0.5 to 0.95 (step size 0.05).

\begin{table}[!t]
    \centering
    \caption{Comparison of different methods under 4-bit weight quantization. ``Time'' denotes quantization times. ``Mem.'' denotes peak memory usage.}
    \label{tab:ResNetw4a32}
    \renewcommand{\arraystretch}{1.3}
    \resizebox{\linewidth}{!}{
      \begin{tabular}{ccccccc}
          \hline % 使用 \hline
          Model & Bit Width & Method & Layer-wise Quant. & Accuracy & Time (s) & Mem. (MB) \\
          \doublemidline
          \multirow{8}[0]{*}{ResNet-18} & FP32 & - & - & 69.76 & - & - \\
          \cdashline{2-7} 
          & \multirow{7}[0]{*}{W4A32}  & \textcolor{gray}{BRECQ} &  & \textcolor{gray}{70.94} & \textcolor{gray}{1789} & \textcolor{gray}{5391} \\
          \cdashline{3-7} 
          &   & Bias Correction & $\checkmark$ & 53.76 & - & - \\
          &   & AdaRound & $\checkmark$ & 68.52 & 1225 & 3834 \\
          &   & AdaQuant & $\checkmark$ & 67.01 & 341 & 4789 \\
          &   & Bit-split & $\checkmark$ & 69.11 & 3191 & 10803 \\
          &   & OBQ & $\checkmark$ & 69.33 & 7784 & 6502 \\
          &   & FastOBQ & $\checkmark$ & \textbf{69.37} & \textbf{58} & \textbf{3069} \\
          \doublemidline
          \multirow{8}[0]{*}{ResNet-50} & FP32 & - & - & 76.13 & - & - \\
          \cdashline{2-7} 
          & \multirow{7}[0]{*}{W4A32}  & \textcolor{gray}{BRECQ} &  & \textcolor{gray}{76.463} & \textcolor{gray}{5558} & \textcolor{gray}{10295} \\
          \cdashline{3-7} 
          &   & Bias Correction & $\checkmark$ & 63.52 & - & - \\
          &   & AdaRound & $\checkmark$ & 75.26 & 3766 & 4517 \\
          &   & AdaQuant & $\checkmark$ & 75.22 & 1127 & 7760 \\
          &   & Bit-split & $\checkmark$ & 75.58 & 5032 & 10856 \\
          &   & OBQ & $\checkmark$ & 75.71 & 9287 & 6859 \\
          &   & FastOBQ & $\checkmark$ & \textbf{75.77} & \textbf{80} & \textbf{3683} \\
          \doublemidline
      \end{tabular}
      }
  \end{table}

\noindent\textbf{Implementation Details}.  
Following the practices of OBQ, the quantization calibration process uses a subset of training data. Specifically, for the ImageNet classification task, 0.1\% of the original training set (approximately 2,048 samples) is randomly sampled to form the calibration dataset, augmented with standard data augmentation techniques (random flipping and cropping) to expand the dataset size by tenfold. For object detection tasks, no data augmentation is applied. Additionally, to ensure numerical stability and invertibility of the Hessian matrix $H$, a damping value of 0.1 is uniformly added to its diagonal elements. After the quantization process, a post-processing step tunes the Batch Normalization (BN) layers in the model. This step adjusts the running statistics (mean and variance) of BN layers to adapt to the activation distribution resulting from quantized weights, mitigating potential negative impacts on model accuracy. The procedure involves resetting the running statistics of all BN layers, performing several forward passes on the quantized model using the calibration dataset, and re-estimating and updating the running mean and variance of the BN layers. During final evaluation, the model uses these tuned BN statistics, which better reflect the data distribution characteristics after quantization.

  \begin{table}[t]
      \centering
      \caption{Comparison of different methods under YOLOv5 model quantization. ``Time'' denotes quantization times. ``Mem.'' denotes peak memory usage.}
      \label{tab:coco_quantres}
      \renewcommand{\arraystretch}{1.3}
      \resizebox{0.97\linewidth}{!}{
        \begin{tabular}{cccccccc}
            \hline % 使用 \hline
            Model & Bit Width & Method & Layer-wise & mAP@0.5 & mAP@0.5:0.95 &  Time (s) & Mem. (MB) \\
            \doublemidline
            \multirow{10}[0]{*}{YOLOv5s} & FP32 & - & - & 37.46 & 56.73 & - & - \\
            \cdashline{2-8} 
            & \multirow{3}[0]{*}{W6A32} & Bias Correction & $\checkmark$ & 26.52 & 45.69 & - & - \\
            &   & OBQ & $\checkmark$ & \textbf{37.01} & \textbf{56.46} & 611 & 1966 \\
            &   & FastOBQ & $\checkmark$ & 36.60 & 56.25 & \textbf{17} & \textbf{1966} \\
            \cline{2-8}
              & FP32 & - & - & 37.46 & 56.73 & - & - \\
              \cdashline{2-8} 
            & \multirow{3}[0]{*}{W8A32} & Bias Correction & $\checkmark$ & 26.95 & 45.89 & - & - \\
            &   & OBQ & $\checkmark$ & \textbf{37.41} & \textbf{56.74} & 766 & 1966 \\
            &   & FastOBQ & $\checkmark$ & 37.36 & 56.70 & \textbf{22} & \textbf{1966} \\
            \doublemidline
            \multirow{10}[0]{*}{YOLOv5m} & FP32 & - & - & 45.16 & 63.88 & - & - \\
            \cdashline{2-8} 
            & \multirow{3}[0]{*}{W6A32} & Bias Correction & $\checkmark$ & 35.33 & 53.96 & - & - \\
            &   & OBQ & $\checkmark$ & \textbf{44.83} & \textbf{63.81} & 3289 & 4350  \\
            &   & FastOBQ & $\checkmark$ & 44.37 & 63.44 & \textbf{36} & \textbf{3007} \\
            \cline{2-8}
              & FP32 & - & - & 45.16 & 63.88 & - & - \\
              \cdashline{2-8} 
            & \multirow{3}[0]{*}{W8A32} & Bias Correction & $\checkmark$ & 35.56 & 53.99 & - & - \\
            &   & OBQ & $\checkmark$ & \textbf{45.09} & 63.87 & 3282 & 4350 \\
            &   & FastOBQ & $\checkmark$ & 45.08 & \textbf{63.88}  & \textbf{39} & \textbf{3007} \\
            \doublemidline
        \end{tabular}
        }
  \end{table}
  \subsection{Comparative Experiments}
  To evaluate the proposed quantization method, experiments compare layer-wise and non-layer-wise approaches on ImageNet and COCO tasks, with results in Tables~\ref{tab:ResNetw4a32}, and \ref{tab:coco_quantres}. On ImageNet, the method matches top methods at 4 bits, outperforming the non-layer-wise BRECQ with significantly lower quantization time (tens to hundreds of times faster) and minimal resource usage. On COCO, at 6-bit and 8-bit settings, it achieves near-identical accuracy to OBQ (0.3\% difference) with hundreds of times faster quantization and reduced resource demands, confirming its effectiveness and efficiency.

\subsection{Ablation Studies}

\textbf{Effectiveness of Sensitivity-Guided Parameter Quantization.} We conduct experiments on three ResNet models of different scales to validate the effectiveness of the proposed sensitivity-guided parameter quantization. The results are shown in Table~\ref{tab:ablation}. To ensure fair comparison, all results based on sensitivity-guided ordering are reproduced under identical settings. For parameter sorting methods in ascending order, the quantized model accuracy significantly decreases, with an average reduction of 0.5\% compared to descending order sorting. Among the various parameter sorting methods tested, the sensitivity-based sorting method achieves a slight accuracy advantage under specific ordering rules, comparable to the heuristic OBQ algorithm. These results indicate that the proposed method, by considering parameter sensitivity and leveraging Taylor expansion-based minimization of layer-wise quantization errors for parameter compensation, better accounts for the importance of parameters to be quantized, thereby enhancing the quantization performance.

\begin{table}[t]
    \centering
    \caption{Accuracy of different parameter quantization orders on ResNet models.}
    \label{tab:ablation}
    \renewcommand{\arraystretch}{1.5}
    \resizebox{\linewidth}{!}{
        \begin{tabular}{cccccccccc}
            \topline
          \multirow{2}{*}{Bits} & \multirow{2}{*}{Method} & \multirow{2}{*}{Parallel} & \multicolumn{3}{c}{ResNet-34} & \multicolumn{3}{c}{ResNet-50} & \multirow{2}{*}{Avg. Acc.} \\
          & & & Accuracy & Time & Memory & Accuracy & Time & Memory & \\
            \doublemidline
            \multirow{10}[0]{*}{W4A32} & err\_des &   & 72.64 & 13564 & 7276 & 75.49 & 9208 & 6859 & 74.07 \\ 
                & err\_asc &   & 72.16 & 13704 & 7276 & 75.32 & 9199 & 6859 & 73.74 \\ 
                & w\_des &   & 72.63 & 13502 & 7276 & 75.57 & 9166 & 6859 & 74.10 \\ 
                & w\_asc &   & 72.04 & 13140 & 7276 & 75.49 & 9280 & 6859 & 73.77 \\ 
                & sensi\_des &   & \textbf{72.83}  & 13511 & 7275 & \textbf{75.58} & 9112 & 6859 & \textbf{74.21} \\ 
                & sensi\_asc &   & 72.24  & 13138 & 7275 & 75.03 & 9263 & 6859 & 73.64 \\ 
            \cline{2-10}
                & No Sorting & $\checkmark$  & 72.65 & 90 & 3842 & 75.52 & 70 & 3683 & 74.09 \\ 
                & err\_des & $\checkmark$  & 73.00 & 100 & 3842 & 75.62 & 79 & 3683 & 74.31 \\ 
                & w\_des & $\checkmark$  & 72.93 & 90 & 3842 & 75.66 & 77 & 3683 & 74.30 \\ 
                & sensi\_des & $\checkmark$  & \textbf{73.12} & 93 & 3842 & \textbf{75.67} & 79 & 3683 & \textbf{74.40} \\ 
            \doublemidline
        \end{tabular}
      }
\end{table}

\noindent\textbf{Effectiveness of Row-Parallel Matrix Parameter Quantization.} As shown in Table~\ref{tab:ablation}, introducing the row-parallel quantization algorithm results in no significant accuracy loss compared to non-parallel quantization experiments across the sorting methods tested. Moreover, significant improvements are observed in quantization time and memory usage. These results demonstrate that, when performing row-parallel parameter quantization under specified rules, the proposed method accounts for the similarity in quantization order across rows guided by specific metrics, enabling efficient row-parallel quantization and substantially reducing the time and resource consumption of quantization.

\section{Conclusions}
We propose FastOBQ, a sensitivity-guided post-training quantization algorithm that enhances efficiency while preserving accuracy. By prioritizing high-sensitivity parameters and using row-parallel quantization with a shared Hessian matrix, FastOBQ reduces quantization error and computational overhead. Experiments on ResNet and YOLOv5 show order-of-magnitude speedups with accuracy comparable to state-of-the-art methods. Our approach is extensible to cross-layer and hybrid-precision quantization, potentially matching retraining-based methods. Future work will explore these extensions, particularly for large-scale models.

\vspace{10pt}

\noindent\textbf{Acknowledgments}.
This work was partially supported by the Joint Funds of the National Natural Science Foundation of China (Grant No.U24A20327), Key-Area Research and Development Program Guangdong Province 2018B010107001,
the Major Key Project of Peng Cheng Laboratory (PCL) PCL2023A08,
Postdoctoral Fellowship Program of CPSF (Grant No.GZC20251043),
and TCL Science and Technology Innovation Fund, China.

\bibliographystyle{splncs04}
\bibliography{mybibliography}

\begin{thebibliography}{10}
\providecommand{\url}[1]{\texttt{#1}}
\providecommand{\urlprefix}{URL }
\providecommand{\doi}[1]{https://doi.org/#1}

\bibitem{banner2019post}
Banner, R., Nahshan, Y., Soudry, D.: Post training 4-bit quantization of convolutional networks for rapid-deployment. Proceedings of the International Conference on Neural Information Processing Systems (NeurIPS)  \textbf{32} (2019)

\bibitem{bengio2013estimating}
Bengio, Y., L{\'e}onard, N., Courville, A.: Estimating or propagating gradients through stochastic neurons for conditional computation. arXiv preprint arXiv:1308.3432  (2013)

\bibitem{brown2020gpt3}
Brown, T.B., Mann, B., Ryder, N., Subbiah, M., Kaplan, J., Dhariwal, P., Neelakantan, A., Shyam, P., Sastry, G., Askell, A., Agarwal, S., Herbert{-}Voss, A., Krueger, G., Henighan, T., Child, R., Ramesh, A., Ziegler, D.M., Wu, J., Winter, C., Hesse, C., Chen, M., Sigler, E., Litwin, M., Gray, S., Chess, B., Clark, J., Berner, C., McCandlish, S., Radford, A., Sutskever, I., Amodei, D.: Language models are few-shot learners. In: Proceedings of the International Conference on Neural Information Processing Systems (NeurIPS) (2020)

\bibitem{choukroun2019low}
Choukroun, Y., Kravchik, E., Yang, F., Kisilev, P.: Low-bit quantization of neural networks for efficient inference. In: Proceedings of the IEEE/CVF International Conference on Computer Vision Workshops (ICCVW). pp. 3009--3018. IEEE (2019)

\bibitem{deng2009imagenet}
Deng, J., Dong, W., Socher, R., Li, L.J., Li, K., Fei-Fei, L.: Imagenet: A large-scale hierarchical image database. In: Proceedings of the IEEE/CVF Conference on Computer Vision and Pattern Recognition (CVPR). pp. 248--255. Ieee (2009)

\bibitem{dong2017learning}
Dong, X., Chen, S., Pan, S.: Learning to prune deep neural networks via layer-wise optimal brain surgeon. Proceedings of the International Conference on Neural Information Processing Systems (NeurIPS)  \textbf{30} (2017)

\bibitem{frantar2022optimal}
Frantar, E., Alistarh, D.: Optimal brain compression: A framework for accurate post-training quantization and pruning. Proceedings of the International Conference on Neural Information Processing Systems (NeurIPS)  \textbf{35},  4475--4488 (2022)

\bibitem{gholami2022survey}
Gholami, A., Kim, S., Dong, Z., Yao, Z., Mahoney, M.W., Keutzer, K.: A survey of quantization methods for efficient neural network inference. In: Low-power computer vision, pp. 291--326. Chapman and Hall/CRC (2022)

\bibitem{gong2019differentiable}
Gong, R., Liu, X., Jiang, S., Li, T., Hu, P., Lin, J., Yu, F., Yan, J.: Differentiable soft quantization: Bridging full-precision and low-bit neural networks. In: Proceedings of the IEEE/CVF International Conference on Computer Vision (ICCV). pp. 4852--4861 (2019)

\bibitem{he2016deep}
He, K., Zhang, X., Ren, S., Sun, J.: Deep residual learning for image recognition. In: Proceedings of the IEEE/CVF Conference on Computer Vision and Pattern Recognition (CVPR). pp. 770--778 (2016)

\bibitem{he2018learning}
He, X., Cheng, J.: Learning compression from limited unlabeled data. In: European Conference on Computer Vision (ECCV). pp. 752--769 (2018)

\bibitem{hubara2021accurate}
Hubara, I., Nahshan, Y., Hanani, Y., Banner, R., Soudry, D.: Accurate post training quantization with small calibration sets. In: Proceedings of the International Conference on Machine Learning (ICML). pp. 4466--4475. PMLR (2021)

\bibitem{jacob2018quantization}
Jacob, B., Kligys, S., Chen, B., Zhu, M., Tang, M., Howard, A., Adam, H., Kalenichenko, D.: Quantization and training of neural networks for efficient integer-arithmetic-only inference. In: Proceedings of the IEEE/CVF Conference on Computer Vision and Pattern Recognition (CVPR). pp. 2704--2713 (2018)

\bibitem{lecun1989optimal}
LeCun, Y., Denker, J., Solla, S.: Optimal brain damage. Proceedings of the International Conference on Neural Information Processing Systems (NeurIPS)  \textbf{2} (1989)

\bibitem{li2021brecq}
Li, Y., Gong, R., Tan, X., Yang, Y., Hu, P., Zhang, Q., Yu, F., Wang, W., Gu, S.: Brecq: Pushing the limit of post-training quantization by block reconstruction. In: Proceedings of the International Conference on Learning Representations (ICLR) (2021)

\bibitem{lin2014microsoft}
Lin, T.Y., Maire, M., Belongie, S., Hays, J., Perona, P., Ramanan, D., Doll{\'a}r, P., Zitnick, C.L.: Microsoft coco: Common objects in context. In: European Conference on Computer Vision (ECCV). pp. 740--755. Springer (2014)

\bibitem{liu2024deepseek}
Liu, A., Feng, B., Xue, B., Wang, B., Wu, B., Lu, C., Zhao, C., Deng, C., Zhang, C., Ruan, C., et~al.: Deepseek-v3 technical report. arXiv preprint arXiv:2412.19437  (2024)

\bibitem{liu2021swin}
Liu, Z., Lin, Y., Cao, Y., Hu, H., Wei, Y., Zhang, Z., Lin, S., Guo, B.: Swin transformer: Hierarchical vision transformer using shifted windows. In: Proceedings of the IEEE/CVF International Conference on Computer Vision (ICCV). pp. 10012--10022 (2021)

\bibitem{liu2023llm}
Liu, Z., Oguz, B., Zhao, C., Chang, E., Stock, P., Mehdad, Y., Shi, Y., Krishnamoorthi, R., Chandra, V.: {LLM-QAT:} data-free quantization aware training for large language models. In: Ku, L., Martins, A., Srikumar, V. (eds.) Findings of the Association for Computational Linguistics (ACL). pp. 467--484. Association for Computational Linguistics (2024)

\bibitem{liu2021post}
Liu, Z., Wang, Y., Han, K., Zhang, W., Ma, S., Gao, W.: Post-training quantization for vision transformer. Proceedings of the International Conference on Neural Information Processing Systems (NeurIPS)  \textbf{34},  28092--28103 (2021)

\bibitem{meller2019same}
Meller, E., Finkelstein, A., Almog, U., Grobman, M.: Same, same but different: Recovering neural network quantization error through weight factorization. In: Proceedings of the International Conference on Machine Learning (ICML). pp. 4486--4495. PMLR (2019)

\bibitem{nagel2020up}
Nagel, M., Amjad, R.A., Van~Baalen, M., Louizos, C., Blankevoort, T.: Up or down? adaptive rounding for post-training quantization. In: Proceedings of the International Conference on Machine Learning (ICML). pp. 7197--7206. PMLR (2020)

\bibitem{nagel2022overcoming}
Nagel, M., Fournarakis, M., Bondarenko, Y., Blankevoort, T.: Overcoming oscillations in quantization-aware training. In: Proceedings of the International Conference on Machine Learning (ICML). pp. 16318--16330. PMLR (2022)

\bibitem{openai2023gpt4}
OpenAI: {GPT-4} technical report. arXiv preprint arXiv:2303.08774  (2023)

\bibitem{touvron2023llama}
Touvron, H., Lavril, T., Izacard, G., Martinet, X., Lachaux, M., Lacroix, T., Rozi{\`{e}}re, B., Goyal, N., Hambro, E., Azhar, F., Rodriguez, A., Joulin, A., Grave, E., Lample, G.: Llama: Open and efficient foundation language models. arXiv preprint arXiv:2302.13971  (2023)

\bibitem{wang2019learning}
Wang, Z., Lu, J., Tao, C., Zhou, J., Tian, Q.: Learning channel-wise interactions for binary convolutional neural networks. In: Proceedings of the IEEE/CVF Conference on Computer Vision and Pattern Recognition (CVPR). pp. 568--577 (2019)

\bibitem{zhang2018lq}
Zhang, D., Yang, J., Ye, D., Hua, G.: Lq-nets: Learned quantization for highly accurate and compact deep neural networks. In: European Conference on Computer Vision (ECCV). pp. 365--382 (2018)

\bibitem{zhao2019improving}
Zhao, R., Hu, Y., Dotzel, J., De~Sa, C., Zhang, Z.: Improving neural network quantization without retraining using outlier channel splitting. In: Proceedings of the International Conference on Machine Learning (ICML). pp. 7543--7552. PMLR (2019)

\end{thebibliography}

\end{document}